\documentclass{article}
\usepackage{ijcai23}

\usepackage{times}
\usepackage{soul}
\usepackage{url}
\usepackage[hidelinks]{hyperref}
\usepackage[utf8]{inputenc}
\usepackage[small]{caption}
\usepackage{graphicx}
\usepackage{amsmath}
\usepackage{amsthm}
\usepackage{booktabs}
\usepackage[switch]{lineno}
\usepackage{xcolor}

\usepackage[ruled, vlined, boxed, linesnumbered]{algorithm2e}
\usepackage{subfigure}
\usepackage{amssymb}

\newcommand{\framework}{SISA}

\title{Hierarchical State Abstraction Based on Structural Information Principles}

\author{
Xianghua Zeng$^{1}$,
Hao Peng$^{1}$,
Angsheng Li$^{1,2}$,
Chunyang Liu$^{3}$,
Lifang He$^{4}$,
Philip S. Yu$^{5}$\\
\affiliations
$^1$ State Key Laboratory of Software Development Environment, Beihang University;\\
$^2$ Zhongguancun Laboratory;\\
$^3$ Didi Chuxing;\\
$^4$ Computer Science \& Engineering, Lehigh University;\\
$^5$ Department of Computer Science, University of Illinois at Chicago.\\
\emails
\{zengxianghua, penghao, angsheng\}@buaa.edu.cn, liangsheng@gmail.zgclab.edu.cn, liuchunyang@didiglobal.com, lih319@lehigh.edu, psyu@uic.edu.
}

\begin{document}
\maketitle
\begin{abstract}
State abstraction optimizes decision-making by ignoring irrelevant environmental information in reinforcement learning with rich observations.
Nevertheless, recent approaches focus on adequate representational capacities resulting in essential information loss, affecting their performances on challenging tasks.
In this article, we propose a novel mathematical \textbf{S}tructural \textbf{I}nformation principles-based \textbf{S}tate \textbf{A}bstraction framework, namely \textbf{\framework}, from the information-theoretic perspective.
Specifically, an unsupervised, adaptive hierarchical state clustering method without requiring manual assistance is presented, and meanwhile, an optimal encoding tree is generated.
On each non-root tree node, a new aggregation function and condition structural entropy are designed to achieve hierarchical state abstraction and compensate for sampling-induced essential information loss in state abstraction.
Empirical evaluations on a visual gridworld domain and six continuous control benchmarks demonstrate that, compared with five SOTA state abstraction approaches, \framework~significantly improves mean episode reward and sample efficiency up to $18.98$ and $44.44\%$, respectively.
Besides, we experimentally show that SISA is a general framework that can be flexibly integrated with different representation-learning objectives to improve their performances further.
\end{abstract}

\section{Introduction}\label{Introduction}
Reinforcement Learning (RL) is a promising approach to intelligent decision-making for a variety of complex tasks, such as robot walking \cite{collins2005efficient}, recommending systems \cite{ijcai2019p360}, automating clustering \cite{zhang2022automating}, abnormal detection \cite{peng2021reinforced}, and multi-agent collaboration \cite{baker2019emergent,peng2022reinforced}.
In the RL setting, agents often learn to maximize their rewards in environments with high-dimensional and noisy observations, which requires suitable state representations \cite{jong2005state,kaiser2019model}.
A valid solution is state abstraction, which can ignore irrelevant environmental information to compress the original state space, thereby considerably simplifying the decision process \cite{abel2016near,laskin2020reinforcement}.

Prior work defines state-abstraction types via aggregation functions that group together ``sufficiently similar" states for reductions in task complexity \cite{li2006towards,hutter2016extreme,abel2016near,abel2018state}.
However, their abstraction performances heavily depend on manual assistance due to high sensitivity to aggregation parameters, such as approximate abstraction’s predicate constant and transitive abstraction’s bucket size.
On the other hand, recent work transfers state abstraction into a representation-learning problem and incorporates various learning objectives to enable state representations with desirable properties \cite{gelada2019deepmdp,zhang2020learning,zang2022simsr,zhu2022invariant}.
Despite their adequate representational capacities, these approaches discard some essential information about state dynamics or rewards, making them hard to characterize the environment accurately.
Therefore, balancing irrelevant and essential information is vital for decision-making with rich observations.
Recently, Markov state abstraction~\cite{allen2021learning} is introduced to realize this balance, reflecting the original rewards and transition dynamics while guaranteeing its representational capacity.
Nevertheless, representation learning based on sampling from finite replay buffers inevitably induces essential information loss in Markov abstraction, affecting its performance on challenging tasks.
Although multi-agent collaborative role discovery based on structural information principles has been proposed \cite{zeng2023effective}, it is not available in the RL scenario of a single agent.

In this paper, we propose a novel mathematical \textbf{S}tructural \textbf{I}nformation principles-based hierarchical \textbf{S}tate \textbf{A}bstraction framework, namely~\framework, from the information-theoretic perspective.
The critical insight is that \framework~combines hierarchical state clustering and aggregation of different hierarchies to achieve sample-efficient hierarchical abstraction.
Inspired by the structural entropy minimization principle~\cite{li2016structural,li2018decoding}, we first present an unsupervised, adaptive hierarchical state clustering method without requiring manual assistance.
It consists of structuralization, sparsification, and optimization modules, to construct an optimal encoding tree.
Secondly, an effective autoencoder structure and representation-learning objectives are adopted to learn state embeddings and refine the hierarchical clustering.
Thirdly, for non-root tree nodes of different heights, we define a new aggregation function using the assigned structural entropy as each child node's weight, thereby achieving the hierarchical state abstraction.
The hierarchical abstraction from leaf nodes to the root on the optimal encoding tree is an automatic process of ignoring irrelevant information and preserving essential information.
Moreover, a new conditional structural entropy is designed to reconstruct the relation between original states to compensate for sampling-induced essential information loss.
Furthermore, \framework~is a general framework and can be flexibly integrated with various representation-learning abstraction approaches, e.g., Markov abstraction \cite{allen2021learning} and SAC-AE \cite{yarats2021improving}, for improving their performances.
Extensive experiments are conducted in both offline and online environments with rich observations, including one gridworld navigation task and six continuous control benchmarks.
Comparative results and analysis demonstrate the performance advantages of the proposed state abstraction framework over the five latest SOTA baselines.
All source codes and experimental results are available at Github\footnote{\url{https://github.com/RingBDStack/SISA}}.

The main contributions of this paper are as follows:
1) Based on the structural information principles, an innovative, unsupervised, adaptive hierarchical state abstraction framework (SISA) without requiring manual assistance is proposed to optimize RL in rich environments.
2) A novel aggregation function leveraging the assigned structural entropy is defined to achieve hierarchical abstraction for efficient decision-making.
3) A new conditional structural entropy reconstructing state relations is designed to compensate for essential information loss in abstraction.
4) The remarkable performance on challenging tasks shows that~\framework~achieves up to $18.98$ and $44.44\%$ improvements in the final performance and sample efficiency than the five latest SOTA baselines.

\section{Background}\label{Background}

\subsection{Markov Decision Process}
In RL, the problem to resolve is described as a Markov Decision Process (MDP) \cite{bellman1957markovian}, a tuple $\mathcal{M}=(\mathcal{S}, \mathcal{A}, \mathcal{R}, \mathcal{P}, \gamma)$, where $\mathcal{S}$ is the original state space, $\mathcal{A}$ is the action space, $\mathcal{R}$ is the reward function, $\mathcal{P}\left(s^{\prime} \mid s, a\right)$ is the transitioning probability from state $\mathrm{s} \in \mathcal{S}$ to state $\mathrm{s}^{\prime} \in \mathcal{S}$ conditioning on an action $a \in \mathcal{A}$, and $\gamma \in[0,1)$ is the discount factor.
At each timestep, an agent chooses an action $a \in \mathcal{A}$ according to its policy function $a \sim \pi(s)$, which updates the environment state $s^{\prime} \sim \mathcal{P}(s, a)$, yielding a reward $r \sim \mathcal{R}(s, a) \in \mathbb{R}$.
The goal of the agent is to learn a policy that maximizes long-term expected discounted reward.
% : $\max _\pi \mathbb{E}_{\mathcal{P}}\left[\sum_{t=0}^{\infty}\left[\gamma^t \mathcal{R}\left(s_t, a_t\right)\right]\right]$.

\subsection{State Abstraction}
% State abstraction effectively reduces the size of high-dimensional or noisy environment state space for tractable decision-making.
Following Markov state abstraction \cite{allen2021learning}, we define state abstraction as a function $f_{\phi}$ that projects each original state $s \in \mathcal{S}$ to an abstract state $z \in \mathcal{Z}$.
When applied to an MDP $\mathcal{M}=(\mathcal{S}, \mathcal{A}, \mathcal{R}, \mathcal{P}, \gamma)$, the state abstraction induces a new abstract decision process $\mathcal{M}_{\phi}=(\mathcal{Z}, \mathcal{A}, \mathcal{R}_{\phi}, \mathcal{P}_{\phi}, \gamma)$, where typically $\left|\mathcal{Z}\right| \ll|\mathcal{S}|$.

\subsection{Structural Information Principles}
Structural information principles were first proposed to measure the dynamical uncertainty of a graph, called structural entropy~\cite{li2016structural}.
They have been widely applied to optimize graph classification and node classification~\cite{wu2022structural,wu2022simple,zou2023se,wang2023user,yang2023minimum}, obfuscate community structures~\cite{liu2019rem}, and decode the chromosomes domains~\cite{li2018decoding}.
By minimizing the structural entropy, we can generate the optimal partitioning tree, which we name an ``encoding tree".
% The encoding tree represents a hierarchical graph partitioning structure.
% Next, we give the formal definitions of the encoding tree, one-dimensional structural entropy, and $K$-dimensional structural entropy.

We suppose a weighted undirected graph $G=(V, E, W)$, where $V$ is the vertex set\footnote{Vertices are defined in the graph, and nodes are in the tree.}, $E$ is the edge set, and $W:E \mapsto \mathbb{R}^{+}$ is the weight function of edges.
Let $n=|V|$ be the number of vertices and $m=|E|$ be the number of edges.
For each graph vertex $v \in V$, the weights sum of its connected edges is defined as its degree $d_{v}$.

\noindent\textbf{Encoding tree.} 
The encoding tree of graph $G$ is a rooted tree defined as follows:
1) For each node $\alpha \in T$, a vertex subset $T_{\alpha}$ in $G$ corresponds with $\alpha$, $T_{\alpha} \subseteq V$.
2) For the root node $\lambda$, we set $T_{\lambda}=V$.
3) For each node $\alpha \in T$, we mark its children nodes as $\alpha^{\wedge}\langle i\rangle$ ordered from left to right as $i$ increases, and ${\alpha^{\wedge}\langle i\rangle}^{-}=\alpha$.
4) For each node $\alpha \in T$, $L$ is supposed as the number of its children; then all vertex subsets $T_{\alpha^{\wedge}\langle i\rangle}$ are disjointed, and $T_{\alpha}=\bigcup_{i=1}^{L} T_{\alpha^{\wedge}\langle i\rangle}$.
5) For each leaf node $\nu$, $T_{\nu}$ is a singleton subset containing a graph vertex.

\noindent\textbf{One-dimensional structural entropy.}
The one-dimensional structural entropy\footnote{It is another form of Shannon entropy using the stationary distribution of the vertex degrees.} measures the dynamical complexity of the graph $G$ without any partitioning structure and is defined as:
\begin{equation}\label{1d_se}
    H^{1}(G)=-\sum_{v \in V} \frac{d_{v}}{vol(G)} \cdot \log _{2} \frac{d_{v}}{vol(G)}\text{,}
\end{equation}
where $vol(G)=\sum_{v \in V} d_{v}$ is the volume of $G$.

\noindent\textbf{$K$-dimensional structural entropy.}
An encoding tree $T$, whose height is at most $K$, can effectively reduce the dynamical complexity of graph $G$, and the $K$-dimensional structural entropy measures the remaining complexity.
For each node $\alpha \in T, \alpha \neq \lambda$, its assigned structural entropy is defined as:
\begin{equation}\label{kd_se_node}
    H^{T}(G;\alpha)=-\frac{g_{\alpha}}{vol(G)} \log _{2} \frac{\mathcal{V}_{\alpha}}{\mathcal{V}_{\alpha^{-}}}\text{,}
\end{equation}
where $g_{\alpha}$ is the sum of weights of all edges connecting vertices in $T_{\alpha}$ with vertices outside $T_{\alpha}$.
$\mathcal{V}_{\alpha}$ is the volume of $T_{\alpha}$, the sum of degrees of vertices in $T_{\alpha}$.
Given the encoding tree $T$, the $K$-dimensional structural entropy of $G$ is defined as:
\begin{equation}\label{kd_se}
    H^{K}(G)=\min_{T}\left\{\sum_{\alpha \in T, \alpha \neq \lambda}H^{T}(G;\alpha)\right\}\text{,}
\end{equation}
where $T$ ranges over all encoding trees whose heights are at most $K$, and the dimension $K$ constraints the maximal height of the encoding tree $T$.

\section{The SISA Framework}\label{State Abstraction}
This section describes the detailed design of the structural information principles-based state abstraction and how to apply SISA to optimize RL.

\begin{figure}[t]
    \centering
    \includegraphics[width=1\columnwidth]{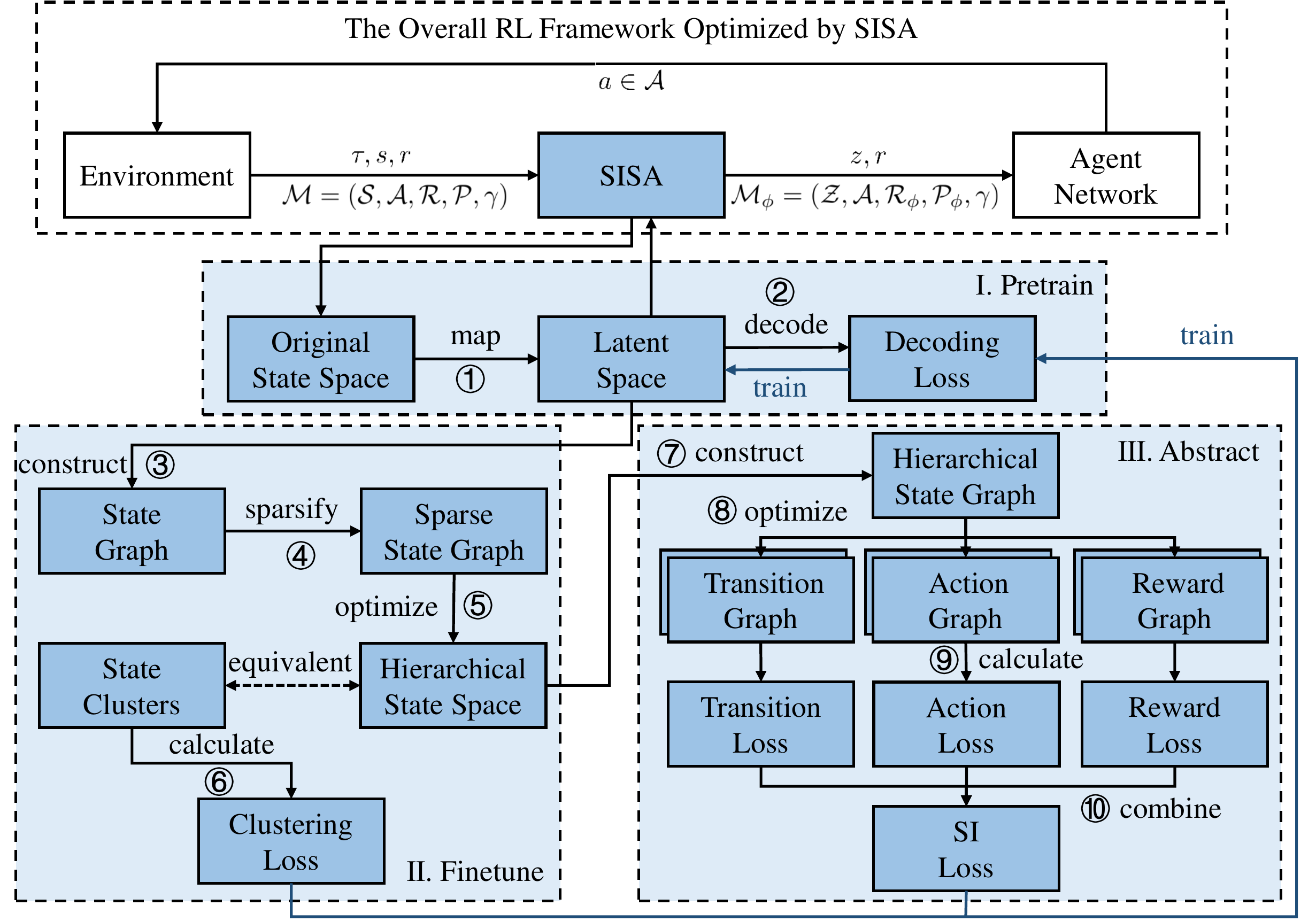}
    \caption{The proposed \framework~framework.}
    \label{overall framework}
\end{figure}

\subsection{Overall RL Framework Optimized by SISA}
For better descriptions, we first introduce how to apply SISA to optimize RL framework.
The optimized RL framework consists of three modules: Environment, Agent Network $\mathcal{Q}$, and the proposed state abstraction \framework, as shown in Fig.~\ref{overall framework}.
The decision process in the environment is labeled as an MDP $\mathcal{M}=(\mathcal{S}, \mathcal{A}, \mathcal{R}, \mathcal{P}, \gamma)$, where the original state space $\mathcal{S}$ is high-dimensional and noisy.
\framework~described in the following subsection takes action-observation history $\tau$ as input and maps each original environment state $s \in \mathcal{S}$ to an abstract state $z \in \mathcal{Z}$, where $\left|\mathcal{Z}\right| \ll|\mathcal{S}|$.
Moreover, the agent makes decisions based on its individual network $\mathcal{Q}$ taking the abstract state $z$ and reward $r$ as inputs, which induces a new abstract decision process $\mathcal{M}_{\phi}=(\mathcal{Z}, \mathcal{A}, \mathcal{R}_{\phi}, \mathcal{P}_{\phi}, \gamma)$.

\subsection{Hierarchical State Abstraction}
As shown in Fig.~\ref{overall framework}, \framework~includes pretrain, finetune, and abstract stages.
In the pretrain stage, we map the original state space to a dense low-dimensional latent space and adopt representation-learning objectives to decode.
In the finetune stage, we sparsify a state graph, optimize its encoding tree to obtain a hierarchical state structure, and calculate a clustering loss.
In the abstract stage, we construct a hierarchical state graph and extract transition, action, and reward relations to calculate a structural information (SI) loss.

\noindent\textbf{Pretrain.}
For tractable decision-making in high-dimensional and noisy environments, we utilize representation-learning objectives to compress the original state space via an abstraction function, as the level-$0$ abstraction.

\begin{figure*}[t]
    \centering
    \includegraphics[width=1\textwidth]{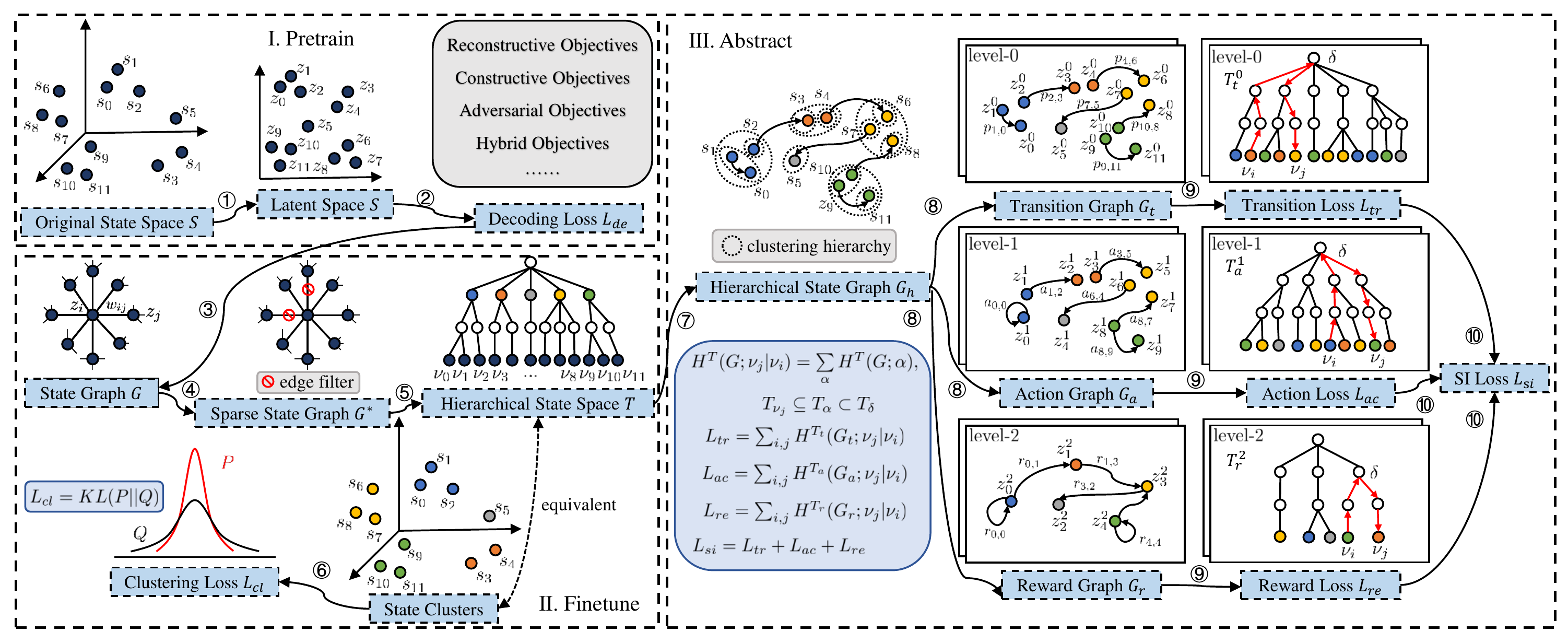}
    \caption{The proposed state abstraction \framework.}
    \label{\framework}
\end{figure*}

To this end, we adopt the encoder-decoder structure \cite{cho2014learning} to learn abstract state representations, mapping the state space $S$ to a low-dimensional and dense abstract state space $Z$.
In the encoder, we encode each state $s \in S$ as a $d$-dimensional embedded representation $z \in Z$ via the abstraction function $f_\phi: S \rightarrow Z$, as the step 1 in Fig.~\ref{\framework}\footnote{For better understanding, we set $S=\{s_{0}, s_{1} \dots, s_{11}\}$ in the original state space as an example.}.
In the decoder, we decode each abstract representation $z$ and select the training objectives in Markov state abstraction, including constructive and adversarial objectives, for calculating the decoding loss $L_{de}$ to guarantee Markov property in the pretrain stage, as the step 2 in Fig.~\ref{\framework}.
Given the action-observation history $\tau$, the encoder-decoder structure is trained end-to-end by minimizing $L_{de}$.  
Furthermore, the abstraction function $f_{\phi}$ will be further optimized in the finetune and abstract stages by minimizing the clustering loss $L_{cl}$ and SI loss $L_{si}$.

\noindent\textbf{Finetune.}
Instead of aggregation condition definitions \cite{abel2018state} or representation learning in the original state space \cite{gelada2019deepmdp,laskin2020curl,zhang2020learning}, we present an unsupervised, adaptive hierarchical clustering method without requiring manual assistance to obtain the hierarchical structure of environment states.
Specifically, we construct a weighted, undirected, complete state graph according to state correlations, minimize its structural entropy to get the optimal encoding tree, and calculate the clustering loss based on Kullback-Leibler (KL) divergence.

Firstly, for states $s_{i}$ and $s_{j}$ with $i \neq j$, we calculate the cosine similarity between abstract representations $z_{i}$ and $z_{j}$ to measure their correlation $\mathcal{C}_{ij} \in\left[-1, 1\right]$.
Intuitively, the larger the value of $\mathcal{C}_{ij}$ represents the more similarity between states $s_{i}$ and $s_{j}$, which should belong to the same cluster with a more significant probability.
We take states as vertices and for any two vertices $s_{i}$ and $s_{j}$, assign $\mathcal{C}_{i,j}$ to the undirected weighted edge $(s_{i}, s_{j})$, $w_{ij} = \mathcal{C}_{ij}$, thereby constructing the complete graph $G$, as the step 3 in Fig.~\ref{\framework}.
In $G$, vertices represent states in $S$, namely $V=S$, edges represent state correlations, and edge weight quantifies the cosine similarity between states.
We define edge weight whose absolute value approaches 0 as trivial weight.

Secondly, we realize sparsification of the state graph to eliminate negative interference of trivial weights.
Following the construction of cancer cell neighbor networks \cite{li2016three}, we minimize the one-dimensional structural entropy to sparsify graph $G$ into a $k$-nearest neighbor ($k$-NN) graph $G_{k}$, as the step 4 in Fig.~\ref{\framework}.
We retain the most significant $k$ edge weights for each vertex to construct $G_{k}$, calculate its one-dimensional structural entropy $H^{1}(G_{k})$, select parameter $k$ of the minimum structural entropy as $k^{*}$, and output $G_{k^{*}}$ as the sparse state graph $G^{*}$.
Moreover, we initialize an encoding tree $T$ of $G^{*}$: 1) We generate a root node $\lambda$ and set its vertex subset $T_{\lambda}=S$ as the whole state space; 2) We generate a leaf node $\nu$ with $T_{\nu}=\{s\}$ for each state $s \in S$, and set it as a child node of $\lambda$, $\nu^{-}=\lambda$.

Thirdly, we realize the hierarchical state clustering by optimizing the encoding tree $T$ from $1$ layer to $K$ layers.
In our work, two operators \textit{merge} and \textit{combine} are introduced from the deDoc \cite{li2018decoding} to optimize the sparse graph $G^{*}$ by minimizing its $K$-dimensional structural entropy, as the step 5 in Fig.~\ref{\framework}.
We define two nodes possessing a common father node in the encoding tree $T$ are brothers.
The \textit{merge} and \textit{combine} are operated on brother nodes and marked as $T_{mg}$ and $T_{cb}$.
We summarize the encoding tree optimization as an iterative algorithm, as shown in Algorithm \ref{alg:optimization}.
At each iteration, we traverse all brother nodes $\beta_{1}$ and $\beta_{2}$ in $T$ (lines 4 and 9) and greedily execute operator $T_{mg}$ or $T_{cb}$ to realize the maximum structural entropy reduction $\Delta SE$ if the tree height does not exceed $K$ (lines 5 and 10).
When no brother nodes satisfy $\Delta SE > 0$ or the tree height exceeds $K$, we terminate the iterative algorithm and output the optimal encoding tree $T^{*}$.
The tree $T^{*}$ is a hierarchical clustering structure of the state space $S$, where the root node $\lambda$ corresponds to $S$, $T_{\lambda}=S$, each leaf node $\nu$ corresponds to a singleton containing a single state $s \in S$, $T_\nu=\{s\}$, and other tree nodes correspond to state clusters with different hierarchies.

Finally, we choose each child $\lambda^{\wedge}\langle i\rangle$ of the root node $\lambda$ as a cluster center and define a structural probability distribution among its corresponding vertex set $T_{\lambda^{\wedge}\langle i\rangle}$ to calculate its embedding $C_{i}$.
For each vertex $s_{j} \in T_{\lambda^{\wedge}\langle i\rangle}$, we define its distribution probability using the sum of the assigned structural entropies of nodes on the path connecting its corresponding leaf node $\nu_{j}$ and node $\lambda^{\wedge}\langle i\rangle$ as follows:
\begin{equation}
    p_{\lambda^{\wedge}\langle i\rangle}(s_{j})=\exp(-\sum\limits_{T_{\nu_{j}} \subseteq T_{\alpha} \subset T_{\lambda^{\wedge}\langle i\rangle}}H^{T^{*}}(G;\alpha))\text{,}
\end{equation}
where $\alpha$ is any node on the path connecting $\nu_{j}$ and $\lambda^{\wedge}\langle i\rangle$.
For the cluster center $\lambda^{\wedge}\langle i\rangle$, we calculate its embedding $C_i$ by:
\begin{equation}
    C_{i}=\sum_{s_{j} \in T_{\lambda^{\wedge}\langle i\rangle}}p_{\lambda^{\wedge}\langle i\rangle}(s_{j}) \cdot z_{j}\text{,}
\end{equation}
where $z_{j}$ is the abstract representation of state $s_{j}$.
Based on the abstract representations and cluster center embeddings, we generate a soft assignment matrix $Q$, where $Q_{ij}$ represents the probability of assigning $i$-th state $s_{i}$ to $j$-th cluster center $\lambda^{\wedge}\langle j\rangle$.
We derive a high-confidence assignment matrix $P$ from $Q$ and calculate the clustering loss $L_{cl}$ as follows:
\begin{equation}\label{clustering loss}
    L_{cl}=KL(P \| Q)=\sum_i \sum_j P_{ij} \log \frac{P_{ij}}{Q_{ij}}\text{.}
\end{equation}
% We will give a detailed introduction to the abstract stage in the following subsection.

\begin{algorithm}[tb]\setcounter{AlgoLine}{0}
    \caption{The Iterative Optimization Algorithm}
    \label{alg:optimization}
    \KwIn{$T$}
    \KwOut{$T^{*}$}
    \begin{small}
        Initialize $\beta_{1}^{*}, \beta_{2}^{*}$
        
        \While{True}{
        $\Delta SE\gets 0$
        
        \For{each brother nodes $\beta_{1}$ and $\beta_{2}$ in $T$}{
        $\beta_{1}^{*}, \beta_{2}^{*}\gets$ maximize $\Delta SE$ caused by the operator $T_{mg}$ via Eq. (\ref{kd_se})
        }
        \If{$\Delta SE > 0$}{
        Execute the operator $T_{mg}$ on $\beta_{1}^{*},\beta_{2}^{*}$
        
        Continue
        }
        \For{each brother nodes $\beta_{1}$ and $\beta_{2}$ in $T$}{
        $\beta_{1}^{*}, \beta_{2}^{*}\gets$ maximize $\Delta SE$ caused by the operator $T_{cb}$ via Eq. (\ref{kd_se})
        }
        \eIf{$\Delta SE > 0$}{
        Execute the operator $T_{cb}$ on $\beta_{1}^{*},\beta_{2}^{*}$
        }
        {
        Break
        }}
        $T^{*}\gets T$
        
        \textbf{return} $T^{*}$
    \end{small}
\end{algorithm}

\begin{figure}[h]
    \centering
    \includegraphics[width=1\columnwidth]{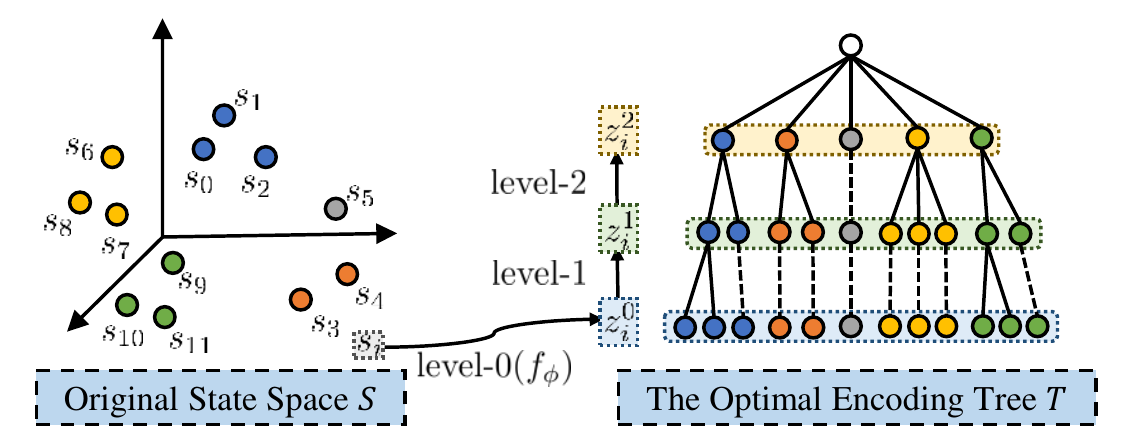}
    \caption{The hierarchical abstraction in the \framework.}
    \label{hierarchical abstractions}
\end{figure}

\subsection{Abstraction on Optimal Encoding Tree}
To compensate for essential information loss induced by sampling, we leverage structural information principles to design an aggregation function on the optimal encoding tree for achieving hierarchical abstraction while accurately characterizing the original decision process.

The optimal encoding tree $T^{*}$ represents a hierarchical clustering structure of the state space $S$, where each tree node corresponds to a state cluster and the height is its clustering hierarchy.
Given the action-observation and reward histories, we firstly sample randomly to construct a hierarchical state graph $G_{h}$, where vertices represent states and edges represent state transitions with action and reward information, as the step 7 in Fig.~\ref{\framework}.
Because of construction by sampling, there is an inevitable essential loss of reward or action information between states in the hierarchical graph $G_{h}$.
Secondly, we define an aggregation function on the optimal encoding tree to achieve hierarchical abstraction from leaf nodes to the root, as shown in Fig.~\ref{hierarchical abstractions}.
For each leaf node $\nu_{i}$ with $T_{\nu_{i}}=\{s_{i}\}$, we define the level-$0$ abstraction via function $f_{\phi}$ described in the pretrain stage and get its level-0 abstract representation $z^{0}_{i}$:
\begin{equation}
    z^{0}_{i} = f_{\phi}(s_{i})\text{.}
\end{equation}
For each non-leaf node $\alpha_{i}$ whose height is $h$, we design an aggregation function using the assigned structural entropy as each child node's weight to achieve the level-$h$ abstraction:
\begin{equation}\label{abstraction function}
        z^{h}_{i} = \sum_{j=1}^{L} \frac{H^{T^{*}}\left(G ; \alpha_i^{\wedge}\langle j\rangle\right)}{\sum_{l=1}^{L}{H^{T^{*}}\left(G ; \alpha_i^{\wedge}\langle l\rangle\right)}} \cdot z^{h-1}_{li+j-1}\text{,}
\end{equation}
where $L$ is the number of children nodes of $\alpha_{i}$ and $li$ is its most left child's index in tree nodes whose height is $h-1$.
Thirdly, we extract three kinds of state relations (transition, action, and reward) from the hierarchical graph $G_{h}$ to construct multi-level transition, action, and reward graphs, respectively, as the step 8 in Fig.~\ref{\framework}.
For convenience, we take the level-$0$ transition graph $G^{0}_{t}$ as an example, and operations on graphs of different relations or levels are similar.
In $G^{0}_{t}$, vertices represent the level-$0$ abstract representations and edge weights quantify the transition probabilities between states via sampling. 
Fourthly, we minimize the $K$-dimensional structural entropy of $G^{0}_{t}$ to generate its optimal encoding tree $T^{0}_{t}$ and calculate the level-$0$ transition loss $L^{0}_{tr}$, as the step 9 in Fig.~\ref{\framework}.
Furthermore, we design a conditional structural entropy to reconstruct the state relation to compensate for sampling-induced essential information loss.
For any two leaf nodes $\nu_{i}$ and $\nu_{j}$ in $T^{0}_{t}$, we find their common father node $\delta$ and calculate conditional structural entropy to quantify the transition probability from $z^{0}_{i}$ to $z^{0}_{j}$ as follows:
\begin{equation}
    p(z^{0}_{j}|z^{0}_{i})=H^{T^{0}_{t}}(G^{0}_{t};z^{0}_{j}|z^{0}_{i})=\sum\limits_{T_{\nu_{j}} \subseteq T_\alpha \subset T_\delta}H^{T^{0}_{t}}(G^{0}_{t};\alpha)\text{,}
\end{equation}
where $\alpha$ is any node on the path connecting the father node $\delta$ and leaf $\nu_{j}$.
And we decode the abstract representations to reconstruct transition probabilities for calculating $L^{0}_{tr}$.
Finally, as the step 10 in Fig.~\ref{\framework}, the SI loss $L_{si}$ is calculated as:
\begin{equation}
    L_{si}=L_{tr}+L_{ac}+L_{re}=\sum_{i=1}^{K}(L^{i}_{tr}+L^{i}_{ac}+L^{i}_{re})\text{,}
\end{equation}
where $K$ is the maximal encoding tree height.

\section{Experiments}\label{Experiments}
In this section, we conduct extensive empirical and comparative experiments, including offline abstraction for visual gridworlds and online abstraction for continuous control.
And we evaluate final performance by measuring the mean reward of each episode and evaluate sample efficiency by measuring how many steps it takes to achieve the best performance.
Similar to other works \cite{laskin2020curl,zhang2020learning}, all experimental results are illustrated with the average and deviation of performances with different random seeds for fair evaluations.
By default, we set the maximal encoding tree height in SISA as 3, $K=3$.
All experiments are conducted on a 3.00GHz Intel Core i9 CPU and an NVIDIA RTX A6000 GPU.

\begin{table*}[t]
\centering
\resizebox{1\linewidth}{!}{
\begin{tabular}{c|cccccc}
\hline
Domain, Task & ball\_in\_cup-catch & cartpole-swingup & cheetah-run & finger-spin & reacher-easy & walker-walk \\ \hline
DBC         &     $168.95 \pm 84.76$         &    $317.74 \pm 77.49$     &    $432.24 \pm 181.43$     &    $805.90 \pm 78.85$    &    $191.44 \pm 69.07$     &    $331.97 \pm 108.40$    \\
SAC-AE      &     $929.24 \pm 39.14$         &    $\underline{839.23} \pm 15.83$     &    $663.71 \pm 9.16$     &    $898.08 \pm 30.23$    &    $917.24 \pm 38.33$     &    $895.33 \pm 56.25$    \\
RAD         &     $\underline{937.97} \pm \textbf{6.77}$          &    $825.62 \pm \underline{9.80}$     &    $\underline{802.53} \pm 8.73$     &    $835.20 \pm 93.26$    &    $908.24 \pm \underline{25.62}$     &    $907.08 \pm 13.02$    \\
CURL        &     $899.03 \pm 30.61$         &    $824.46 \pm 18.53$     &    $309.49 \pm \textbf{8.15}$     &    $949.57 \pm 15.71$    &    $\underline{919.71} \pm 28.03$     &    $885.03 \pm \textbf{9.88}$    \\
Markov      &     $919.10 \pm 38.14$         &    $814.94 \pm 17.61$     &    $642.79 \pm 65.92$     &    $\underline{969.91} \pm \textbf{8.41}$    &     $806.34 \pm 131.40$    &    $\underline{918.44} \pm 12.58$    \\
\hline
\framework (Ours)   &     $\textbf{946.29} \pm \underline{8.63}$          &    $\textbf{858.21} \pm \textbf{6.31}$      &    $\textbf{806.67} \pm \underline{8.61}$     &    $\textbf{970.45} \pm \underline{8.75}$    &    $\textbf{924.52} \pm \textbf{19.04}$     &    $\textbf{921.64} \pm \underline{12.43}$    \\
\hline
% \multicolumn{7}{c}{Improvements}                          \\ \hline
Abs.($\%$) Avg. $\uparrow$     &       $8.32(0.89)$        &    $18.98(2.26)$     &     $4.14(0.52)$    &   $0.54(0.06)$     &    $4.81(0.52)$     &    $3.20(0.35)$    \\
\hline
\end{tabular}}
\caption{Summary of the mean episode rewards for different tasks from DMControl: ``average value $\pm$ standard deviation" and ``average improvement" (absolute value($\%$)). \textbf{Bold}: the best performance under each category, \underline{underline}: the second performance.}
\label{table}
\end{table*}

\subsection{Offline Abstraction for Visual Gridworlds}
\noindent\textbf{Experimental setup.}
First, we evaluate SISA for offline state abstraction in a visual gridworld domain, where each discrete position is mapped to a noisy image, like experiments in Markov abstraction \cite{allen2021learning}.
The agent only has access to these noisy images and uses a uniform random exploration policy over four directional actions to train the SISA framework offline.
Then, we froze the framework that maps images to abstract states while training DQN \cite{mnih2015human}.
We compare SISA against three baselines, including pixel prediction \cite{kaiser2019model}, reconstruction \cite{lee2020stochastic}, and Markov abstraction \cite{allen2021learning}.

\begin{figure}[h]
    \centering
    \subfigure[]{
        \label{msa_gridworld}
        \centering
        \includegraphics[width=0.515\linewidth]{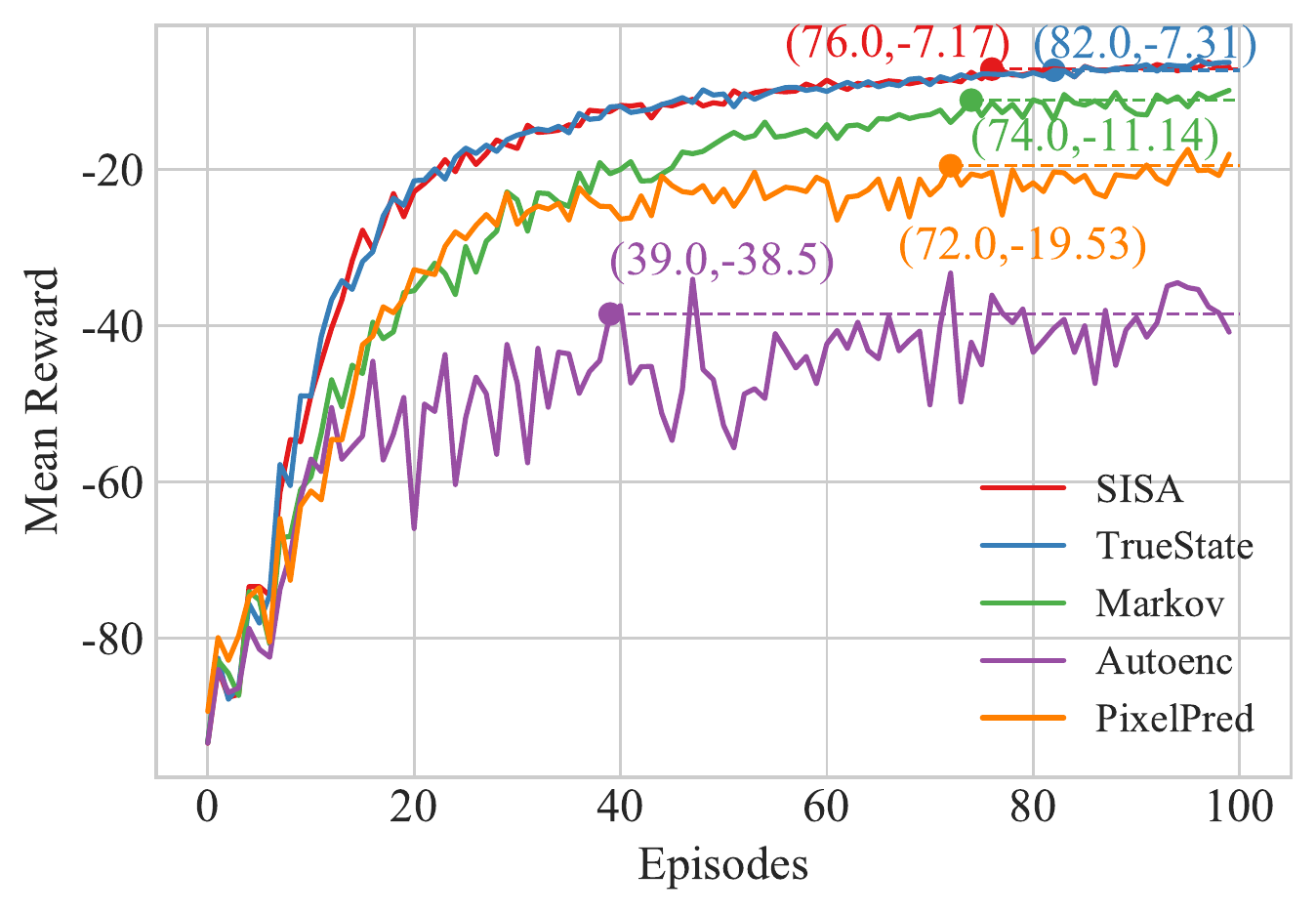}
    }
    \subfigure[]{
        \label{visual state abstraction}
        \centering
        \includegraphics[width=0.425\linewidth]{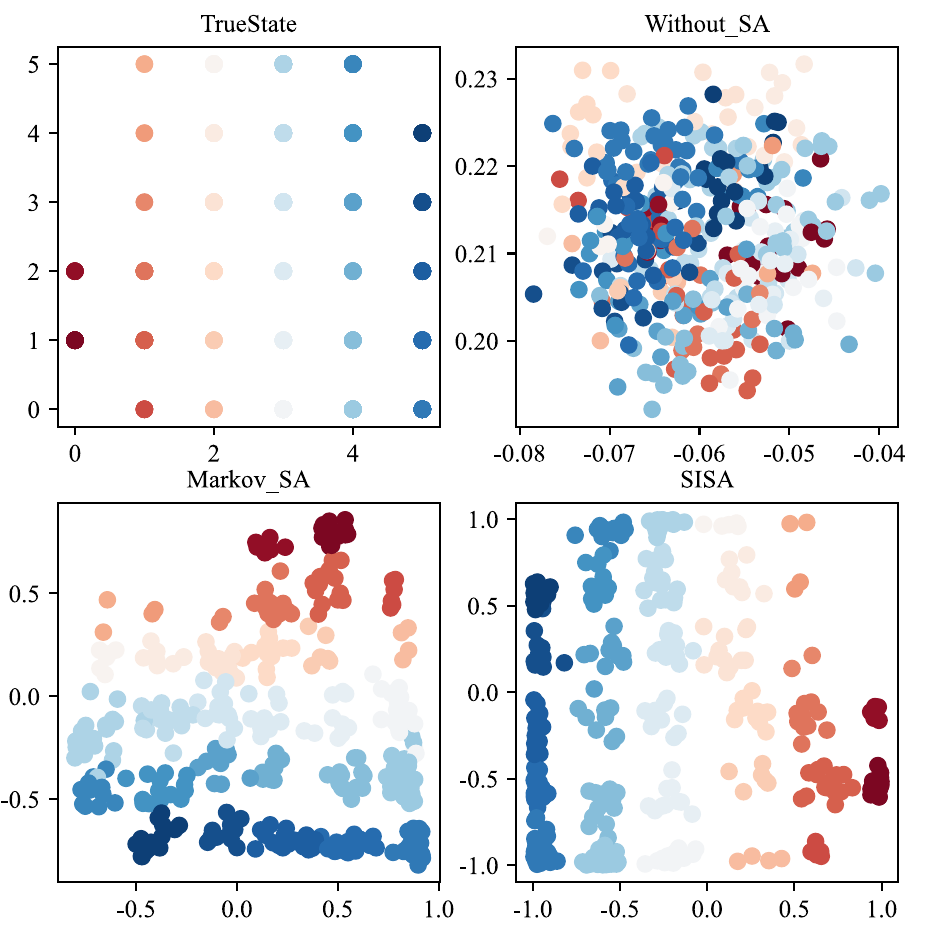}
    }
    \caption{(a) Mean episode rewards for the visual gridworld navigation task. (b) Visualization of 2-D state abstractions for the $6 \times 6$ visual gridworld domain.}
\end{figure}

\noindent\textbf{Evaluations.}
Fig.~\ref{msa_gridworld} shows the learning curves of SISA and three baselines for the visual gridworld navigation task.
For reference, we also include a learning curve for DQN trained on ground-truth positions without abstraction, labeled as TrueState.
Each curve's starting point of convergence is marked in brackets.
As shown in Fig.~\ref{msa_gridworld}, SISA converges at $76.0$ epochs and achieves a $-7.17$ mean episode reward.
It can be observed that SISA significantly outperforms other baselines and matches the performance of the TrueState.
Moreover, we visualize the 2-D abstract representations for the $6 \times 6$ gridworld domain and denote ground-truth positions with different colors in Fig.~\ref{visual state abstraction}.
In \framework, the hierarchical clustering based on the structural information principles effectively reconstructs relative positions of the gridworld better than baselines, resulting in its advantage.

\subsection{Online Abstraction for Continuous Control}
\noindent\textbf{Experimental setup.}
Next, we benchmark our framework in an online setting with a challenging and diverse set of image-based, continuous control tasks from the DeepMind Control suite (DMControl) \cite{tunyasuvunakool2020dm_control}.
The online experiments are conducted on six DMControl environments: $ball\_in\_cup$-$catch$, $cartpole$-$swingup$, $cheetah$-$run$, $finger$-$spin$, $reacher$-$easy$, and $walker$-$walk$, to examine the sample efficiency and final performance.
The Soft Actor-Critic (SAC) \cite{haarnoja2018soft} is chosen as a traditional RL algorithm, combined with SISA and different state abstraction baselines.
The compared state-of-the-art baselines consist of random data augmentation RAD \cite{laskin2020reinforcement}, contrastive method CURL \cite{laskin2020curl}, bisimulation method DBC \cite{zhang2020learning}, pixel-reconstruction method SAC-AE \cite{yarats2021improving}, and Markov abstraction \cite{allen2021learning}.

\noindent\textbf{Evaluations.}
We evaluate all compared methods in six environments from the DMControl suite and summarize averages and deviations of mean episode rewards in Table \ref{table}.
It can be seen that SISA improves the average mean episode reward in each DMControl environment.
Specifically, SISA achieves up to $18.98$ $(2.26\%)$ improvement from $839.23$ to $858.21$ in average value, which corresponds to its advantage on final performance.
In terms of stability, SISA reduces the standard deviation in two environments.
And in the other four environments, SISA achieves the second lowest deviations ($8.63$, $8.61$, $8.75$, and $12.43$), where it remains very close to the best baseline.
The reason is that, SISA minimizes the structural entropy to realize the optimal hierarchical state clustering without any manual assistance and therefore guarantees its stability.

\begin{figure}[t]
    \centering
    \includegraphics[width=1\columnwidth]{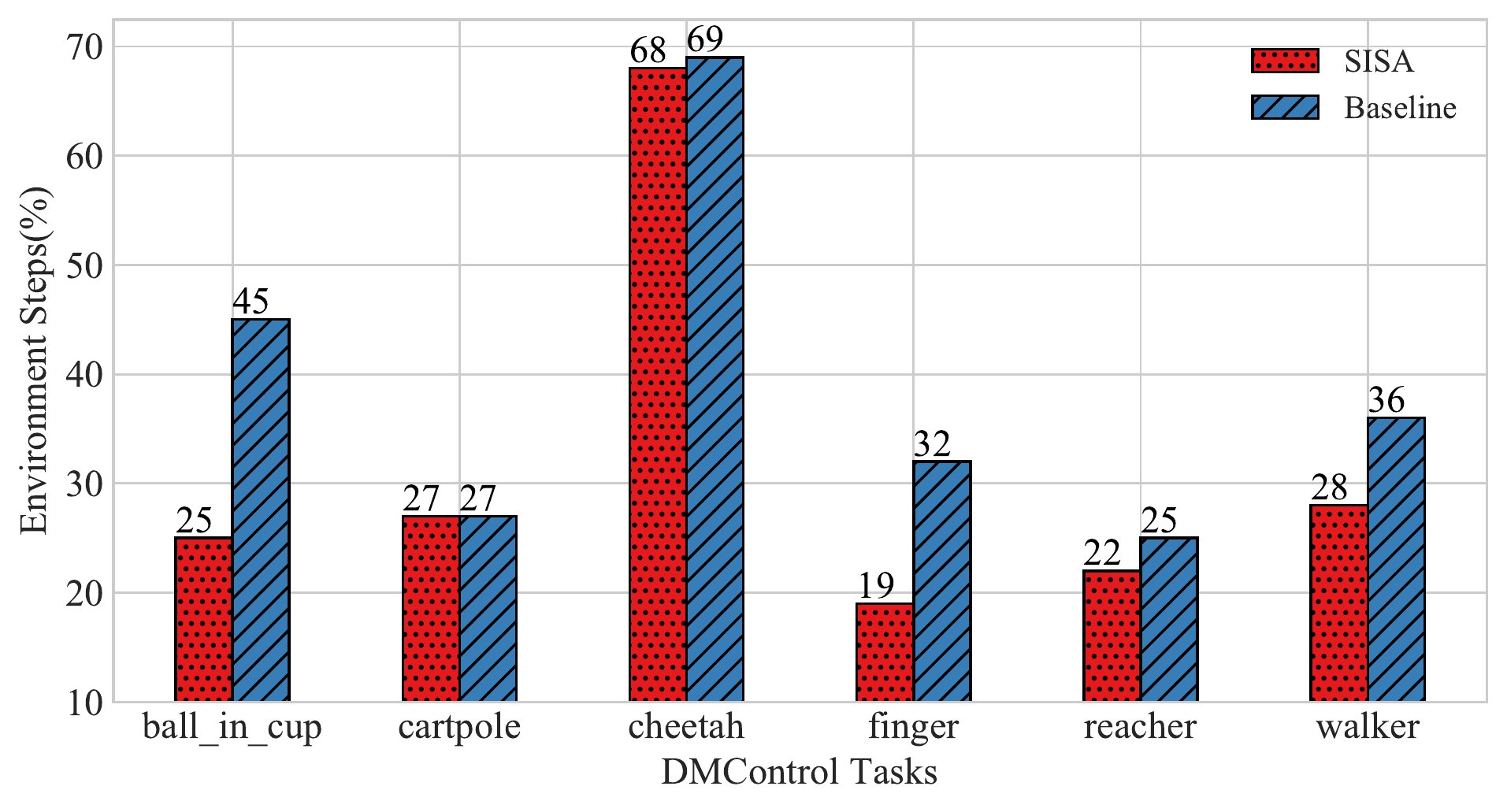}
    \caption{The sample-efficiency results for DMControl experiments.}
    \label{sample-efficiency}
\end{figure}

On the other hand, the sample-efficiency results of DMControl experiments are shown in Fig.~\ref{sample-efficiency}.
In each experiment, we set the mean reward target as 0.9 times the final performance of SISA and choose the best baseline as the compared method.
In contrast to classical baselines, SISA takes fewer steps to finish the mean episode reward target and thereby achieves higher sample efficiency.
In particular, SISA achieves up to $44.44\%$ improvement in sample efficiency, reducing the environment steps from $45k$ to $25k$ to obtain an $851.661$ mean reward in $ball\_in\_cup$-$catch$ task.

\begin{figure*}[t]
    \centering
    \includegraphics[width=1\textwidth]{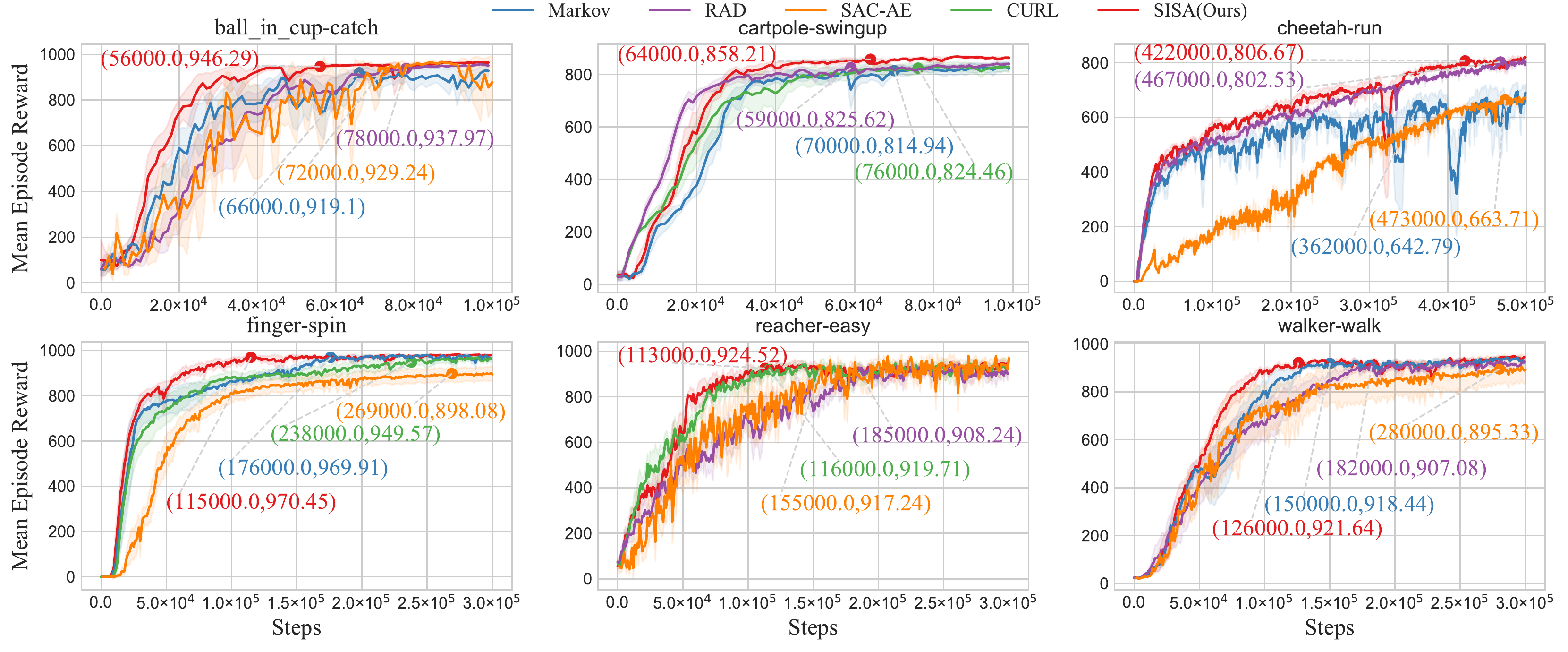}
    \caption{Mean episode rewards on six DMControl environments.}
    \label{msa_dmcontrol}
\end{figure*}

In summary, in the online setting where reward information is available, SISA establishes a new state of art on DMControl regarding final performance and sample efficiency.
The hierarchical abstraction on the optimal encoding tree effectively compensates for essential information loss in state compression to maintain the original task characteristics, guaranteeing \framework's advantages.
Fig.~\ref{msa_dmcontrol} shows the learning curves of SISA and the three best-performing baselines in each task; similarly, their starting points of convergence are marked.
SISA converges at $64000.0$ timesteps and achieves an $858.21$ mean episode reward, as shown in the $cartpole$-$swingup$ task.

\subsection{Integrative Abilities}
\begin{figure}[t]
    \centering
    \subfigure[ball\_in\_cup-catch]{
        \centering
        \includegraphics[width=1\linewidth]{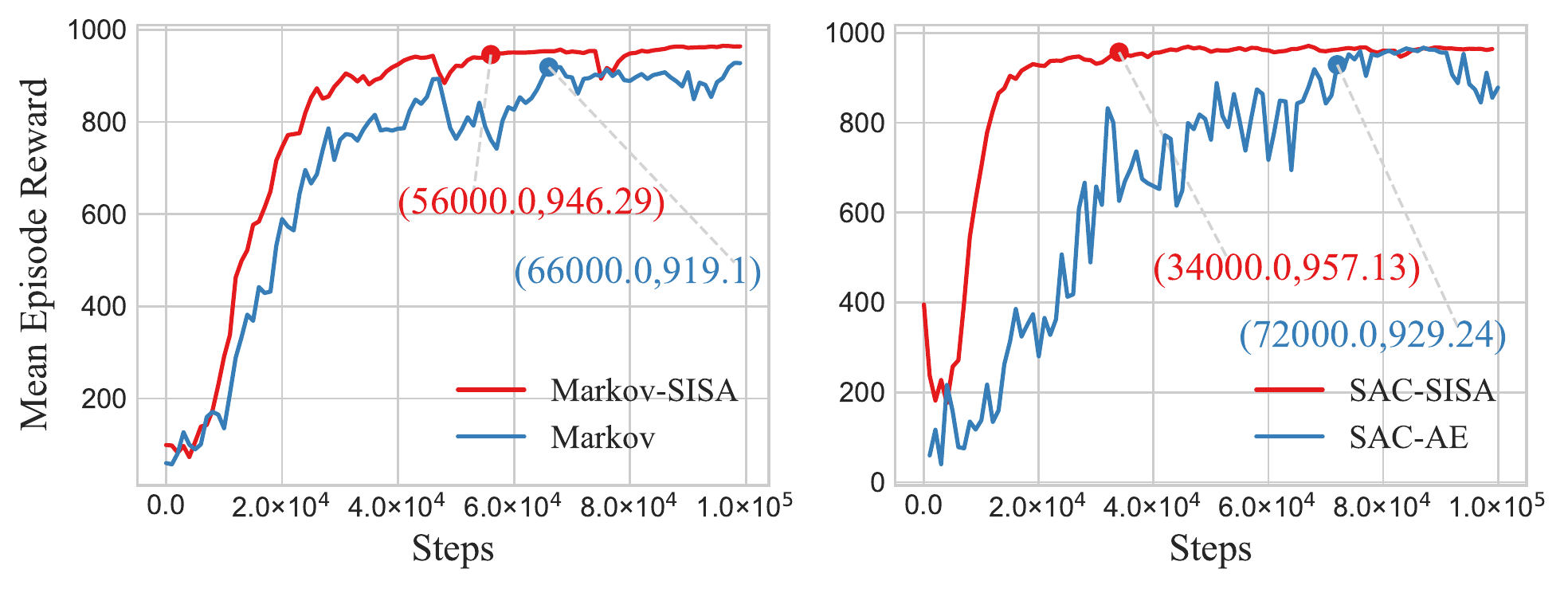}
    }
    \subfigure[cartpole-swingup]{
        \centering
        \includegraphics[width=1\linewidth]{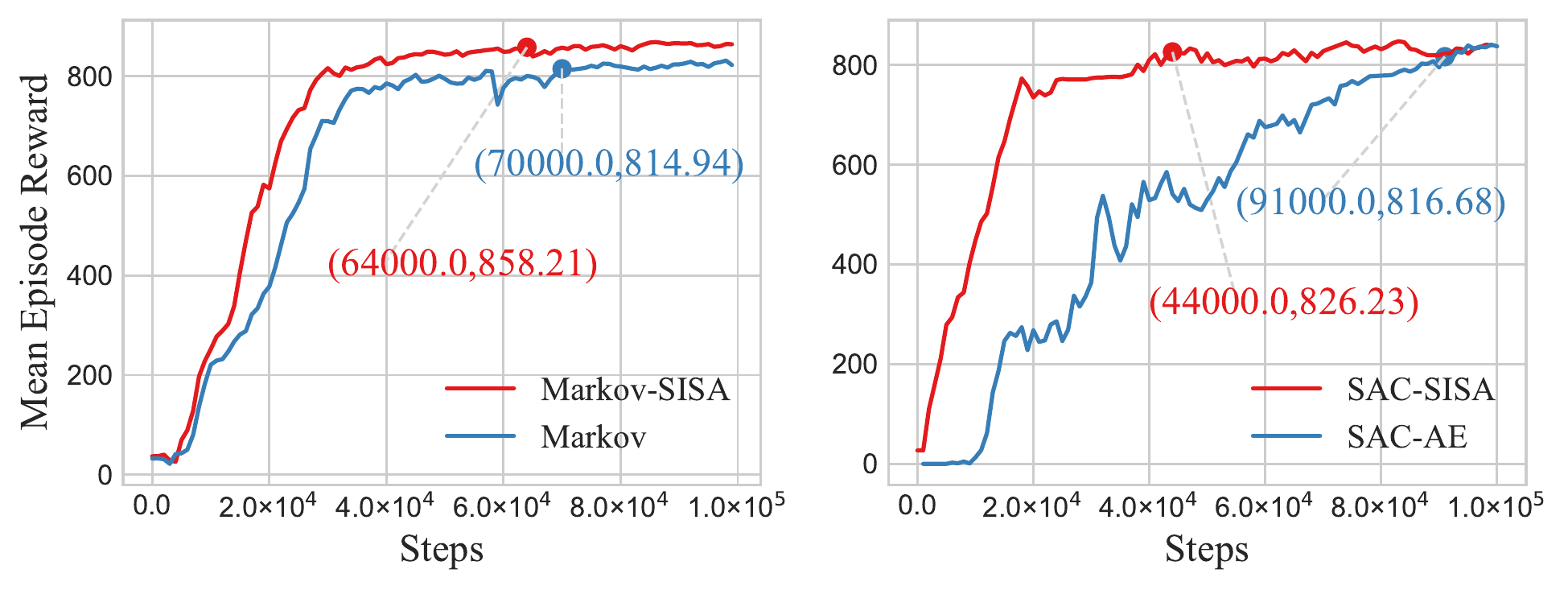}
    }
    \caption{Mean episode rewards of the SISA integrated with abstraction methods Markov and SAC-AE.}
    \label{integrate}
\end{figure}

SISA is a general framework and can be flexibly integrated with various existing representation-learning abstraction approaches in the pretrain stage.
Therefore, we integrate our framework with the Markov abstraction and SAC-AE, namely Markov-\framework~and SAC-\framework, and choose two tasks ($ball\_in\_cup$-$catch$ and $cartpole$-$swingup$) to evaluate their performances.
Each integrated framework achieves higher final performance and sample efficiency than the original approach, as shown in Fig.~\ref{integrate}. 
The experimental results indicate that our abstraction framework can significantly optimize existing abstraction approaches in complex decision-making.

\subsection{Ablation Studies}
\begin{figure}[t]
    \centering
    \includegraphics[width=1\columnwidth]{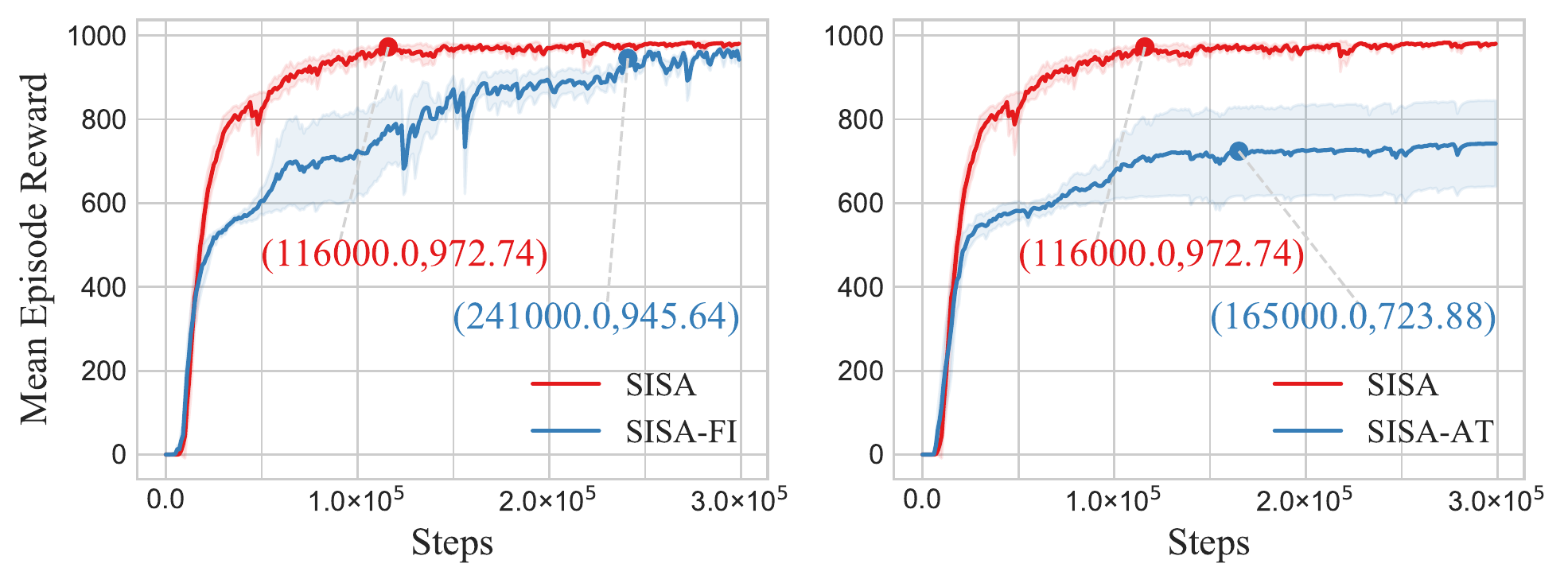}
    \caption{Mean episode rewards for ablation studies.}
    \label{ablation}
\end{figure}

We conduct ablation studies in the $finger$-$spin$ task to understand the functionalities of finetune and abstract stages in \framework.
The finetune and abstract stages are removed from \framework, respectively, and we name the corresponding variants \framework-FI and \framework-AT.
In the abstract stage, \framework-FI optimizes the state graph to obtain the optimal encoding tree of the environment states as the foundation of hierarchical abstraction.
As shown in Fig.~\ref{ablation}, SISA remarkably outperforms \framework-FI and \framework-AT in the final performance, sample efficiency, and stability, which shows that the finetune and abstract stages are both important for the \framework's advantages.
Furthermore, the more significant advantages over the \framework-AT variant indicate that the hierarchical abstraction in the abstract stage is an indispensable key module of SISA.

\section{Related Work}\label{Related Work}
\noindent\textbf{State abstractions for sample-efficient RL.}
% Due to space limitations, we only briefly introduce related work on state abstraction in this subsection.
The SAC-AE \cite{yarats2021improving} trains models to reproduce original states by pixel prediction and related tasks perfectly.
Instead of prediction, the CURL \cite{laskin2020curl} learns abstraction by differentiating whether two augmented views come from the same observation.
The DBC \cite{zhang2020learning} trains a transition model and reward function end-to-end to learn approximate bisimulation abstractions, where original states are equivalent if their expected reward and transition dynamics are the same.
To ensure the abstract decision process is Markov, Allen et al. \shortcite{allen2021learning} introduce sufficient conditions to learn Markov abstract state representations.
Recently, SimSR \cite{zang2022simsr} designs a stochastic approximation method to learn abstraction from observations to robust latent representations.
IAEM \cite{zhu2022invariant} efficiently obtains abstract representations, by capturing action invariance.

\section{Conclusion}\label{Conclusion}
This paper proposes a general structural information principles-based hierarchical state abstraction (\framework) framework, from the information-theoretic perspective.
To the best of our knowledge, it is the first work to incorporate the mathematical structural information principles into state abstraction to optimize decision-making with high-dimension and noisy observations.
Evaluations of challenging tasks in the visual gridworld and DMControl suite demonstrate that \framework~significantly improves final performance and sample efficiency over state-of-the-art baselines.
In the future, we will evaluate \framework~in other environments and further explore the hierarchical encoding tree structure in decision-making.

%% The file named.bst is a bibliography style file for BibTeX 0.99c

% \section*{Acknowledgments}
% The corresponding authors are Hao Peng and Angsheng Li. 
% This paper was supported by the National Key R\&D Program of China through grant 2021YFB1714800, NSFC through grant 61932002, S\&T Program of Hebei through grant 20310101D, Natural Science Foundation of Beijing Municipality through grant 4222030, CCF-DiDi GAIA Collaborative Research Funds for Young Scholars, the Fundamental Research Funds for the Central Universities, Xiaomi Young Scholar Funds for Beihang University, and in part by NSF under grants III-1763325, III-1909323,  III-2106758, and SaTC-1930941. 

\bibliographystyle{named}
\bibliography{0-main}

% Generated by IEEEtran.bst, version: 1.14 (2015/08/26)
\begin{thebibliography}{10}
\providecommand{\url}[1]{#1}
\csname url@samestyle\endcsname
\providecommand{\newblock}{\relax}
\providecommand{\bibinfo}[2]{#2}
\providecommand{\BIBentrySTDinterwordspacing}{\spaceskip=0pt\relax}
\providecommand{\BIBentryALTinterwordstretchfactor}{4}
\providecommand{\BIBentryALTinterwordspacing}{\spaceskip=\fontdimen2\font plus
\BIBentryALTinterwordstretchfactor\fontdimen3\font minus
  \fontdimen4\font\relax}
\providecommand{\BIBforeignlanguage}[2]{{%
\expandafter\ifx\csname l@#1\endcsname\relax
\typeout{** WARNING: IEEEtran.bst: No hyphenation pattern has been}%
\typeout{** loaded for the language `#1'. Using the pattern for}%
\typeout{** the default language instead.}%
\else
\language=\csname l@#1\endcsname
\fi
#2}}
\providecommand{\BIBdecl}{\relax}
\BIBdecl

\bibitem{zeng2023effective}
X.~Zeng, H.~Peng, and A.~Li, ``Effective and stable role-based multi-agent
  collaboration by structural information principles,'' in \emph{Proceedings of
  the AAAI Conference on Artificial Intelligence}, 2023.

\bibitem{collins2005efficient}
S.~Collins, A.~Ruina, R.~Tedrake, and M.~Wisse, ``Efficient bipedal robots
  based on passive-dynamic walkers,'' \emph{Science}, vol. 307, no. 5712, pp.
  1082--1085, 2005.

\bibitem{ijcai2019p360}
E.~Ie, V.~Jain, J.~Wang, S.~Narvekar, R.~Agarwal, R.~Wu, H.-T. Cheng,
  T.~Chandra, and C.~Boutilier, ``Slateq: A tractable decomposition for
  reinforcement learning with recommendation sets,'' in \emph{Proceedings of
  the International Joint Conference on Artificial Intelligence}.\hskip 1em
  plus 0.5em minus 0.4em\relax International Joint Conferences on Artificial
  Intelligence Organization, 2019, pp. 2592--2599.

\bibitem{zhang2011coordinated}
C.~Zhang and V.~R. Lesser, ``Coordinated multi-agent reinforcement learning in
  networked distributed pomdps,'' in \emph{Proceedings of the AAAI Conference
  on Artificial Intelligence}, 2011, pp. 764--770.

\bibitem{baker2019emergent}
B.~Baker, I.~Kanitscheider, T.~M. Markov, Y.~Wu, G.~Powell, B.~McGrew, and
  I.~Mordatch, ``Emergent tool use from multi-agent autocurricula,'' in
  \emph{Proceedings of the International Conference on Learning
  Representations}, 2020, pp. 1--28.

\bibitem{sutton1998introduction}
R.~S. Sutton, A.~G. Barto \emph{et~al.}, ``Introduction to reinforcement
  learning,'' 1998.

\bibitem{dietterich1999state}
T.~Dietterich, ``State abstraction in maxq hierarchical reinforcement
  learning,'' \emph{Advances in Neural Information Processing Systems},
  vol.~12, 1999.

\bibitem{li2006towards}
L.~Li, T.~J. Walsh, and M.~L. Littman, ``Towards a unified theory of state
  abstraction for mdps,'' in \emph{AI\&M}, 2006.

\bibitem{andre2002state}
D.~Andre and S.~J. Russell, ``State abstraction for programmable reinforcement
  learning agents,'' in \emph{Proceedings of the AAAI Conference on Artificial
  Intelligence}, 2002, pp. 119--125.

\bibitem{jong2005state}
N.~K. Jong and P.~Stone, ``State abstraction discovery from irrelevant state
  variables.'' in \emph{Proceedings of the International Joint Conference on
  Artificial Intelligence}, vol.~8.\hskip 1em plus 0.5em minus 0.4em\relax
  Citeseer, 2005, pp. 752--757.

\bibitem{abel2016near}
D.~Abel, D.~Hershkowitz, and M.~Littman, ``Near optimal behavior via
  approximate state abstraction,'' in \emph{Proceedings of the International
  Conference on Machine Learning}.\hskip 1em plus 0.5em minus 0.4em\relax PMLR,
  2016, pp. 2915--2923.

\bibitem{hutter2016extreme}
M.~Hutter, ``Extreme state aggregation beyond markov decision processes,''
  \emph{Theoretical Computer Science}, vol. 650, pp. 73--91, 2016.

\bibitem{abel2018state}
D.~Abel, D.~Arumugam, L.~Lehnert, and M.~Littman, ``State abstractions for
  lifelong reinforcement learning,'' in \emph{Proceedings of the International
  Conference on Machine Learning}.\hskip 1em plus 0.5em minus 0.4em\relax PMLR,
  2018, pp. 10--19.

\bibitem{gelada2019deepmdp}
C.~Gelada, S.~Kumar, J.~Buckman, O.~Nachum, and M.~G. Bellemare, ``Deepmdp:
  Learning continuous latent space models for representation learning,'' in
  \emph{Proceedings of the International Conference on Machine Learning}.\hskip
  1em plus 0.5em minus 0.4em\relax PMLR, 2019, pp. 2170--2179.

\bibitem{laskin2020curl}
M.~Laskin, A.~Srinivas, and P.~Abbeel, ``Curl: Contrastive unsupervised
  representations for reinforcement learning,'' in \emph{Proceedings of the
  International Conference on Machine Learning}.\hskip 1em plus 0.5em minus
  0.4em\relax PMLR, 2020, pp. 5639--5650.

\bibitem{zhang2020learning}
A.~Zhang, R.~T. McAllister, R.~Calandra, Y.~Gal, and S.~Levine, ``Learning
  invariant representations for reinforcement learning without
  reconstruction,'' in \emph{Proceedings of the International Conference on
  Learning Representations}, 2020.

\bibitem{allen2021learning}
C.~Allen, N.~Parikh, O.~Gottesman, and G.~Konidaris, ``Learning markov state
  abstractions for deep reinforcement learning,'' \emph{Advances in Neural
  Information Processing Systems}, vol.~34, pp. 8229--8241, 2021.

\bibitem{abel2019state}
D.~Abel, D.~Arumugam, K.~Asadi, Y.~Jinnai, M.~L. Littman, and L.~L. Wong,
  ``State abstraction as compression in apprenticeship learning,'' in
  \emph{Proceedings of the AAAI Conference on Artificial Intelligence},
  vol.~33, no.~01, 2019, pp. 3134--3142.

\bibitem{shannon1953lattice}
C.~Shannon, ``The lattice theory of information,'' \emph{Transactions of the
  IRE Professional Group on Information Theory}, vol.~1, no.~1, pp. 105--107,
  1953.

\bibitem{li2016structural}
A.~Li and Y.~Pan, ``Structural information and dynamical complexity of
  networks,'' \emph{IEEE Transactions on Information Theory}, vol.~62, no.~6,
  pp. 3290--3339, 2016.

\bibitem{li2016three}
A.~Li, X.~Yin, and Y.~Pan, ``Three-dimensional gene map of cancer cell types:
  Structural entropy minimisation principle for defining tumour subtypes,''
  \emph{Scientific Reports}, vol.~6, pp. 1--26, 2016.

\bibitem{li2018decoding}
A.~Li, X.~Yin, B.~Xu, D.~Wang, J.~Han, Y.~Wei, Y.~Deng, Y.~Xiong, and Z.~Zhang,
  ``Decoding topologically associating domains with ultra-low resolution hi-c
  data by graph structural entropy,'' \emph{Nature Communications}, vol.~9, pp.
  1--12, 2018.

\bibitem{bellman1957markovian}
R.~Bellman, ``A markovian decision process,'' \emph{Journal of Mathematics and
  Mechanics}, pp. 679--684, 1957.

\bibitem{cho2014learning}
K.~Cho, B.~V. Merri{\"e}nboer, C.~Gulcehre, D.~Bahdanau, F.~Bougares,
  H.~Schwenk, and Y.~Bengio, ``Learning phrase representations using rnn
  encoder-decoder for statistical machine translation,'' in \emph{Proceedings
  of the Conference on Empirical Methods in Natural Language Processing}, 2014,
  pp. 1724--1734.

\bibitem{mnih2015human}
V.~Mnih, K.~Kavukcuoglu, D.~Silver, A.~A. Rusu, J.~Veness, M.~G. Bellemare,
  A.~Graves, M.~Riedmiller, A.~K. Fidjeland, G.~Ostrovski \emph{et~al.},
  ``Human-level control through deep reinforcement learning,'' \emph{Nature},
  vol. 518, no. 7540, pp. 529--533, 2015.

\bibitem{kaiser2019model}
{\L}.~Kaiser, M.~Babaeizadeh, P.~Mi{\l}os, B.~Osi{\'n}ski, R.~H. Campbell,
  K.~Czechowski, D.~Erhan, C.~Finn, P.~Kozakowski, S.~Levine \emph{et~al.},
  ``Model based reinforcement learning for atari,'' in \emph{Proceedings of the
  International Conference on Learning Representations}, 2019.

\bibitem{lee2020stochastic}
A.~X. Lee, A.~Nagabandi, P.~Abbeel, and S.~Levine, ``Stochastic latent
  actor-critic: Deep reinforcement learning with a latent variable model,''
  \emph{Advances in Neural Information Processing Systems}, vol.~33, pp.
  741--752, 2020.

\bibitem{tunyasuvunakool2020dm_control}
S.~Tunyasuvunakool, A.~Muldal, Y.~Doron, S.~Liu, S.~Bohez, J.~Merel, T.~Erez,
  T.~Lillicrap, N.~Heess, and Y.~Tassa, ``dm\_control: Software and tasks for
  continuous control,'' \emph{Software Impacts}, vol.~6, p. 100022, 2020.

\bibitem{haarnoja2018soft}
T.~Haarnoja, A.~Zhou, P.~Abbeel, and S.~Levine, ``Soft actor-critic: Off-policy
  maximum entropy deep reinforcement learning with a stochastic actor,'' in
  \emph{Proceedings of the International Conference on Machine Learning}.\hskip
  1em plus 0.5em minus 0.4em\relax PMLR, 2018, pp. 1861--1870.

\bibitem{laskin2020reinforcement}
M.~Laskin, K.~Lee, A.~Stooke, L.~Pinto, P.~Abbeel, and A.~Srinivas,
  ``Reinforcement learning with augmented data,'' \emph{Advances in Neural
  Information Processing Systems}, vol.~33, pp. 19\,884--19\,895, 2020.

\bibitem{yarats2021improving}
D.~Yarats, A.~Zhang, I.~Kostrikov, B.~Amos, J.~Pineau, and R.~Fergus,
  ``Improving sample efficiency in model-free reinforcement learning from
  images,'' in \emph{Proceedings of the AAAI Conference on Artificial
  Intelligence}, vol.~35, no.~12, 2021, pp. 10\,674--10\,681.

\bibitem{liu2019rem}
Y.~Liu, J.~Liu, Z.~Zhang, L.~Zhu, and A.~Li, ``Rem: From structural entropy to
  community structure deception,'' \emph{Advances in Neural Information
  Processing Systems}, vol.~32, 2019.

\bibitem{wu2022structural}
J.~Wu, X.~Chen, K.~Xu, and S.~Li, ``Structural entropy guided graph
  hierarchical pooling,'' in \emph{Proceedings of the International Conference
  on Machine Learning}.\hskip 1em plus 0.5em minus 0.4em\relax PMLR, 2022, pp.
  24\,017--24\,030.

\bibitem{wu2022simple}
J.~Wu, S.~Li, J.~Li, Y.~Pan, and K.~Xu, ``A simple yet effective method for
  graph classification,'' \emph{Proceedings of the International Joint
  Conference on Artificial Intelligence}, 2022.

\bibitem{zang2022simsr}
H.~Zang, X.~Li, and M.~Wang, ``Simsr: Simple distance-based state
  representations for deep reinforcement learning,'' in \emph{Proceedings of
  the AAAI Conference on Artificial Intelligence}, vol.~36, no.~8, 2022, pp.
  8997--9005.

\bibitem{zhu2022invariant}
Z.-M. Zhu, S.~Jiang, Y.-R. Liu, Y.~Yu, and K.~Zhang, ``Invariant action effect
  model for reinforcement learning,'' in \emph{Proceedings of the AAAI
  Conference on Artificial Intelligence}, vol.~36, no.~8, 2022, pp. 9260--9268.

\bibitem{zou2023se}
D.~Zou, H.~Peng, X.~Huang, R.~Yang, J.~Li, J.~Wu, C.~Liu, and P.~S. Yu,
  ``Se-gsl: A general and effective graph structure learning framework through
  structural entropy optimization,'' in \emph{Proceedings of the Web
  Conference}, 2023, pp. 1--12.

\bibitem{wang2023user}
Y.~Wang, Y.~Wang, Z.~Zhang, S.~Yang, K.~Zhao, and J.~Liu, ``User: Unsupervised
  structural entropy-based robust graph neural network,'' in \emph{Proceedings
  of the AAAI Conference on Artificial Intelligence}, 2023.

\bibitem{yang2023minimum}
Z.~Yang, G.~Zhang, J.~Wu, J.~Yang, Q.~Z. Sheng, H.~Peng, A.~Li, S.~Xue, and
  J.~Su, ``Minimum entropy principle guided graph neural networks,'' in
  \emph{Proceedings of the Sixteenth ACM International Conference on Web Search
  and Data Mining}, 2023, pp. 114--122.

\bibitem{zhang2022automating}
R.~Zhang, H.~Peng, Y.~Dou, J.~Wu, Q.~Sun, Y.~Li, J.~Zhang, and P.~S. Yu,
  ``Automating dbscan via deep reinforcement learning,'' in \emph{Proceedings
  of the 31st ACM International Conference on Information \& Knowledge
  Management}, 2022, pp. 2620--2630.

\bibitem{peng2022reinforced}
H.~Peng, R.~Zhang, S.~Li, Y.~Cao, S.~Pan, and P.~S. Yu, ``Reinforced,
  incremental and cross-lingual event detection from social messages,''
  \emph{IEEE Transactions on Pattern Analysis and Machine Intelligence},
  vol.~45, no.~1, pp. 980--998, 2022.

\bibitem{peng2021reinforced}
H.~Peng, R.~Zhang, Y.~Dou, R.~Yang, J.~Zhang, and P.~S. Yu, ``Reinforced
  neighborhood selection guided multi-relational graph neural networks,''
  \emph{ACM Transactions on Information Systems (TOIS)}, vol.~40, no.~4, pp.
  1--46, 2021.

\end{thebibliography}

\end{document}